\documentclass[runningheads]{llncs}
\usepackage{graphicx}
\usepackage{mathtools}
\usepackage{xcolor}      
\usepackage{graphicx}
\usepackage{bm}
\usepackage{makecell}
\usepackage{bbm}
\usepackage{xspace}
\usepackage{float}
\usepackage{color}
\usepackage{colortbl}
\usepackage{dsfont}
\usepackage{multirow}
\usepackage{multicol}
\usepackage{pgfplots}
\usepackage{algorithm}
\usepackage{algpseudocode}
\usepackage{adjustbox}
\usepackage[figuresright]{rotating}
\usepackage{listings}
\usepackage{bbding}
\usepackage{hyperref}
\usepackage{pifont}
\usepackage{wasysym}
\usepackage{amssymb}
\usepackage[misc]{ifsym}

\begin{document}
\title{ICDAR 2023 Competition on Reading the Seal Title}

\author{Wenwen Yu\inst{1} \and
Mingyu Liu\inst{1} \and Mingrui Chen\inst{1} \and Ning Lu\inst{2}  \and Yinlong Wen\inst{3}  \and Yuliang Liu\inst{1}  \and Dimosthenis Karatzas\inst{4} \and  Xiang Bai\inst{1}\textsuperscript{(\Letter)}%
}

\authorrunning{W. Yu et al.}
\institute{Huazhong University of Science and Technology, China\\
\email{\{wenwenyu, mingyuliu, charmier, ylliu, xbai\}@hust.edu.cn}\\ \and
Huawei Technologies Ltd., China\\
\email{luning12@huawei.com}\\ \and
Sichuan Optical Character Technology Co., Ltd.\\
\email{ylwen@chineseocr.com}
\and
Computer Vision Centre, Universitat Autónoma de Barcelona, Spain\\
\email{dimos@cvc.uab.es}}
\maketitle              %
\begin{abstract}
Reading seal title text is a challenging task due to the variable shapes of seals, curved text, background noise, and overlapped text. However, this important element is commonly found in official and financial scenarios, and has not received the attention it deserves in the field of OCR technology. To promote research in this area, we organized ICDAR 2023 competition on reading the seal title (ReST), which included two tasks: seal title text detection (Task 1) and end-to-end seal title recognition (Task 2). We constructed a dataset of 10,000 real seal data, covering the most common classes of seals, and labeled all seal title texts with text polygons and text contents. The competition opened on 30th December, 2022 and closed on 20th March, 2023. The competition attracted 53 participants and received 135 submissions from academia and industry, including 28 participants and 72 submissions for Task 1, and 25 participants and 63 submissions for Task 2, which demonstrated significant interest in this challenging task. In this report, we present an overview of the competition, including the organization, challenges, and results. We describe the dataset and tasks, and summarize the submissions and evaluation results. The results show that significant progress has been made in the field of seal title text reading, and we hope that this competition will inspire further research and development in this important area of OCR technology.
\end{abstract}
\section{Introduction}
\begin{figure*}[ht]
 \centering
 \includegraphics[width=0.8\textwidth]{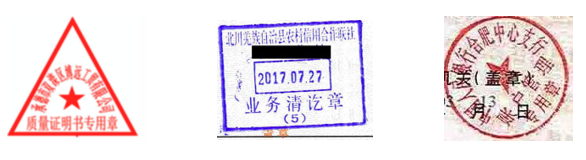}
\caption{Different shapes of seals samples in the ReST.}
\label{fig:rest_examples_shapes}
\end{figure*}

\begin{figure*}[ht]
 \centering
 \includegraphics[width=0.8\textwidth]{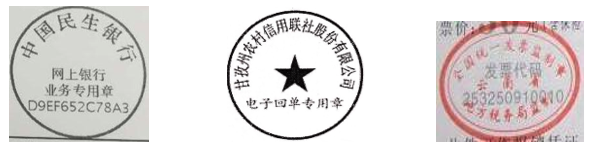}
\caption{Seals with curved texts in the ReST.}
\label{fig:rest_examples_curved}
\end{figure*}

\begin{figure*}[ht]
 \centering
 \includegraphics[width=0.8\textwidth]{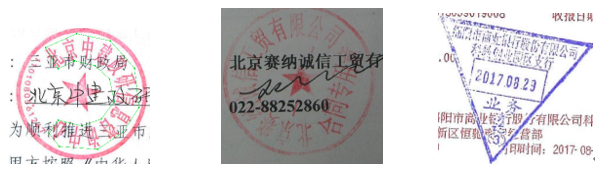}
\caption{Seals with overlapped texts in the ReST.}
\label{fig:rest_examples_overlap}
\end{figure*}

Based on the flourish of deep learning method, we have witnessed the maturity of regular and general OCR technology, including scene text detection and recognition. However, as a common element which can be seen everywhere in official and financial scenarios, seal title text has not gain its attention. And the task of reading seal title text is also faced with many challenges, such as variable shapes of seal (for example, circle, ellipse, triangle and rectangle), curved text, background noise and overlapped text, as shown in Figure~\ref{fig:rest_examples_shapes}~-~\ref{fig:rest_examples_overlap}. In order to promote the research of seal text, we propose the competition on reading the seal title.

Considering there are no existing datasets for seal title text reading. We construct a dataset including 10,000 real seal data, which covers the most common classes of seal. In the dataset, all seal title texts are labeled with text polygons and text contents. Besides, two tasks are presents for this competition: (1) Seal title text detection; (2) End-to-end seal title recognition. We hope that the dataset and tasks could greatly promote the research in seal text reading.

\subsection{Competition Organization}
ICDAR 2023 competition on reading the seal title is organized by a joint team, including Huazhong University of Science and Technology and Universitat Autónoma de Barcelona.

We organize the competition on the Robust Reading Competition (RRC) website~\footnote{\url{https://rrc.cvc.uab.es/?ch=20}}, where provide corresponding download links of the datasets, and user interfaces for participants and submission page for their results.  Great support has been received from the RRC web team. The online evaluation server~\footnote{\url{https://rrc.cvc.uab.es/?ch=20\&com=mymethods\&task=1}} will remain available for future usage of this benchmark.

\section{Dataset and Annotations}
We name our dataset ReST, as it focuses on Reading Seal Title text. It totally includes 10,000 images collected from real scene. The data is mainly in Chinese, with English data accounting for 1\%.

The datasets cover the most common classes of seals:

\begin{itemize}
    \item  \textbf{Circle/Ellipse shapes}: This type of seals are commonly existing in official seals, invoice seals, contract seals, and bank seals.
    \item \textbf{Rectangle shapes}: This type of seals are commonly seen in driving licenses, corporate seals, and medical bills. 
    \item \textbf{Triangle shapes}: This type of seals are seen in bank receipts and financial occasions. This type is uncommon seal and has a small amount of data.
\end{itemize}

The dataset is split half into a training set and a test set. Every image in the dataset is annotated with text line locations and the labels. Locations are annotated in terms of polygons, which are in clockwise order. Transcripts are UTF-8 encoded strings. Annotations for an image are stored in a json file with the identical file name, following the naming convention: gt\_[image\_id], where image\_id refers to the index of the image in the dataset.

In the JSON file, each gt\_[image\_id] corresponds to a list, where each line in the list correspond to one text instance in the image and gives its bounding box coordinates and transcription, in the following format:

\{

“gt\_1”: [

  {“points”: [[x1, y1], [x2, y2], …, [xn, yn]], “transcription” : “trans1” }],

“gt\_2”: [

  {“points”: [[x1, y1], [x2, y2], …, [xn, yn]] , “transcription” : “trans3” }],
……
\}

\noindent where x1, y1, x2, y2, …, xn, yn in “points” are the coordinates of the polygon bounding boxes,. The “transcription” denotes the text of each text line.

Note: There may be some inaccurate annotations in the training set, which can measure the robustness of the algorithm, and participants may filter this part of the data as appropriate. The test set is manually corrected and the annotations are accurate.

\section{Competition Tasks and Evaluation Protocols}
The competition include two tasks: 1) seal title text detection, where the objective is to localize the title text in seal image. and 2) the end-to-end seal title recognition, where the main objective of this task is to extract the title of a seal.

\subsection{Task 1: Seal Title Text detection}
\begin{figure}[ht]
 \centering
 \includegraphics[width=0.98\textwidth]{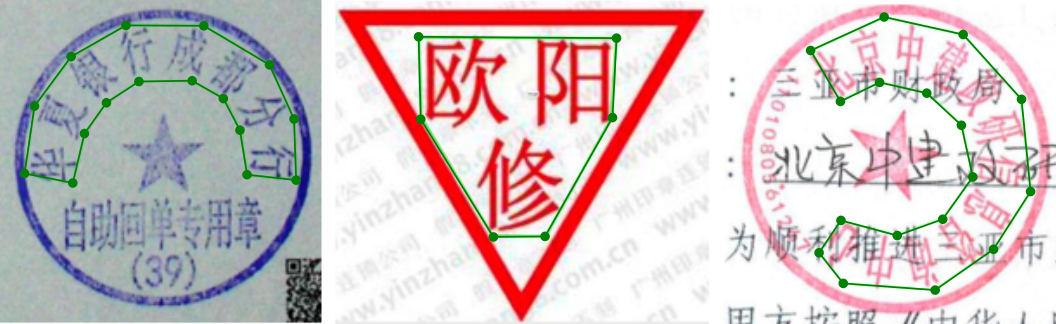}
\caption{Example images of the Seal Text dataset. Green color binding lines are formed with polygon ground truth format.}
\label{fig:rest_task1_demo}
\end{figure}

The aim of this task is to localize the title text in seal image, The input examples are shown in Figure~\ref{fig:rest_task1_demo}.

\textbf{Submission Format.} Participants will be asked to submit a JSON file containing results for all test images. The results format is:

\{

“res\_1”: [

  {“points”: [[x1, y1], [x2, y2], …, [xn, yn]], “confidence” : c}],

“res\_2”: [

  {“points”: [[x1, y1], [x2, y2], …, [xn, yn]] , “confidence” : c }],

……

\}

\noindent where the key of JSON file should adhere to the format of res\_[image\_id]. Also, n is the total number of vertices (could be unfixed, varied among different predicted text instance), and c is the confidence score of the prediction and the range is 0-1.

\textbf{Evaluation Protocol.} For Task 1, we adopt IoU-based evaluation protocol by following CTW1500~\cite{liu2019curved,chng2019icdar2019}. IoU is a threshold-based evaluation protocol, with 0.5 set as the default threshold. We will report results on 0.5 and 0.7 thresholds but only H-Mean under 0.7 will be treated as the final score for each submitted model, and to be used as submission ranking purpose. To ensure fairness, the competitors are required to submit confidence score for each detection, and thus we can iterate all confidence thresholds to find the best H-Mean score. Meanwhile, in the case of multiple matches, we only consider the detection region with the highest IOU, the rest of the matches will be counted as False Positive. The calculation of Precision, Recall, and F-score are as follows:
\begin{equation}
\begin{gathered}
\text { Precision }=\frac{T P}{T P+F P}, \\
\text { Recall }=\frac{T P}{T P+F N}, \\
F=\frac{2 * \text { Precision } * \text { Recall }}{\text { Precision }+ \text { Recall }}
\end{gathered}
\end{equation}

\noindent where TP, FP, FN and F denote true positive, false positive, false negative and H-Mean, respectively.

\subsection{Task 2: End-to-end Seal Title Recognition.}
\begin{figure}[ht]
 \centering
 \includegraphics[width=0.98\textwidth]{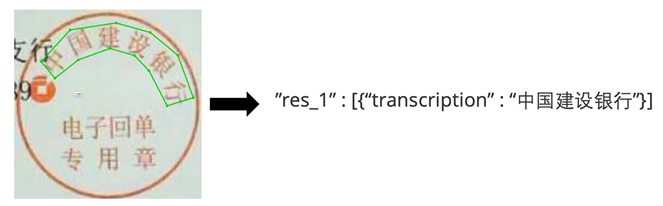}
\caption{ Example of the task2 input-output.}
\label{fig:rest_task2_demo}
\end{figure}

The main objective of this task is to extract the title of a seal, as shown in Figure~\ref{fig:rest_task2_demo}, the input is a whole seal image and the output is the seal’s title.

\textbf{Submission Format.} For Task 2, participants are required to submit the predicted titles for all the images in a single JSON file:

\{

“res\_1”: [{ “transcription” : “title1”}],

“res\_2”: [{ “transcription” : “title2”}],

“res\_3”: [{ “transcription” : “title3”}],

……

\}

\noindent where the key of JSON file should adhere to the format of res\_[image\_id].

\textbf{Evaluation Protocol.} Metrics for this task is case-insensitive word accuracy. We will compute the ratio of correctly predicted titles and the total titles.

\section{Submissions and Results}
By the submission deadline, we received 135 submission from 53 participants in total, including 72 submissions from 28 participants for Task 1, and 63 submissions from 25 participants for Task 2. 

After the submission deadlines, we collected all submissions and evaluate their performance through automated process with scripts developed by the RRC web team. Participants did not receive feedback during the submission process, and for those who made multiple submissions, only the last submission prior to the final deadline was considered for ranking purposes. The winners are determined for each task based on the score achieved by the corresponding primary metric. The complete leaderboard can be accessed on the official competition website~\footnote{\url{https://rrc.cvc.uab.es/?ch=20\&com=evaluation\&task=1}} for all tasks. The following table presents the top 10 results due to limited space.

\subsection{Task 1 Seal Title Text Detection}
The result for Task 1 is presented on Table~\ref{tab:task1_results}.

\begin{table*}[htbp]
\centering
\caption{Task-1: Seal Title Text Detection Results.}
\scalebox{0.71}{
\begin{tabular}{lllllll}
\hline
Rank & Method Name & Team Members & Insititute & Precision-0.7 & Recall-0.7 & Hmean-0.7 \\
\hline
1    & \makecell[l]{Dao Xianghu\\ light of\\ TianQuan} & \makecell[l]{Kai Yang, Ye Wang,\\ Bin Wang, Wentao Liu,\\ Xiaolu Ding, Jun Zhu,\\ Ming Chen, Peng Yao, \\Zhixin Qiu}                                                                 & \makecell[l]{CCB \\Financial \\Technology \\Co. Ltd, China}                                                              & 99.06\%       & 99.06\%    & 99.06\%   \\
\hline
2    & det314\_4                     & Huajian Zhou                                                                                                                                                   & China Mobile Cloud Centre                                                                            & 98.18\%       & 98.18\%    & 98.18\%   \\
\hline
3    & INTIME\_OCR                   &\makecell[l] {Wei Wang, Chengxiang Ran,\\ Jin Wei, Xinye Yang,\\ Tianjiao Cao, Fangmin Zhao}                                                                                      & \makecell[l]{Institute of \\Information \\Engineering,\\ Chinese Academy\\ of Sciences; \\Mashang Consumer \\Finance Co., Ltd }& 98.14\%       & 98.06\%    & 98.10\%   \\
\hline
4    & AntFin-UperNet                & Yangkun Lin, Tao Xu                                                                                                                                            & Ant Group                                                                                            & 97.72\%       & 97.70\%    & 97.71\%   \\
\hline
5    & SPDB LAB                      & \makecell[l]{Jie Li, Wei Wang,\\Yuqi Zhang, Ruixue Zhang,\\Yiru Zhao, Danya Zhou,\\Di Wang, Dong Xiang,\\Hui Wang, Min Xu,\\Pengyu Chen, Bin Zhang,\\Chao Li, Shiyu Hu,\\Songtao Li, Yunxin Yang} & \makecell[l]{Shanghai \\Pudong \\Development Bank}                     & 97.60\%       & 97.60\%    & 97.60\%   \\
\hline
6    & Aaaaa\_v3                     & Wudao, Liaoming                                                                                                                                                & cmb                                                                                                  & 97.34\%       & 97.32\%    & 97.33\%   \\
\hline
7    & PAN\_ReST\_4                  & \makecell[l]{Yuchen Su, Yongkun Du,\\ Tianlun Zheng, \\Yi Gan, Zhineng Chen}                                                                                       & \makecell[l]{Fudan University,\\ Paddle OCR}                                                                         & 96.86\%       & 96.86\%    & 96.86\%   \\
\hline
8    & DB with SegFormer             & Sehwan Joo, Wonho Song                                                                                                                                         & Upstage AI                                                                                           & 98.11\%       & 95.42\%    & 96.75\%   \\
\hline
9    & AppAI for Seal                & \makecell[l]{Chuanjian Liu, Miao Rang,\\ Zhenni Bi, Zhicheng Liu,\\ Wenhui Dong, Yuyang Li, \\Dehua Zheng, Hailin Wu,\\ Kai Han, Yunhe Wang}                                         & Noah                                                                                                 & 96.00\%       & 96.00\%    & 96.00\%   \\
\hline
10   & ratio\_4.0 & sunyifan& SY\_007& 95.96\%       & 95.96\%    & 95.96\%     \\
\hline
\end{tabular}}
\label{tab:task1_results}
\end{table*}

The methods used by the top 3 submissions for Task 1 are
presented below.

\textbf{1st ranking method.}   
The team of ``CCB Financial Technology Co. Ltd, China" are elaborated in detail from the following three perspectives:
\begin{itemize}
    \item Data Analysis:
        This competition provided 5000 pieces of training data officially. Upon analyzing the data, they found that it can be classified into four categories: round, oval, square, and triangular, with the round and oval categories being the primary ones. The training set contains various conditions, including multi-directional rotations, uneven colors, overlapping seals, and indistinct seal patterns.
    \item Data Processing:
        When it comes to data analysis, they began by re-annotating the training set images and enlarging them to squares. They then rotated the data and produced a total of 15,000 images. Data generation was carried out on difficult samples, including those with overlapping or blurry stamps. Prior to generating the seal data, they gathered a large number of company and organization names from the internet. Then, they generated the rotation angle and position of each individual character based on its length and merged them into the seal's background image. Moreover, they output the coordinates of the outer edge points of the text. To create a more realistic representation of seals in the generated data, they incorporated various colors, fonts, backgrounds, and textures. The base image for each seal was created by randomly cropping backgrounds, and they used RGBA format during data generation to allow for control over the color depth of the seal by adding a transparency channel. They also included two types of seal borders: solid and fragmented.
    \item Model Introduction:
        In this segmentation task, they employed a ``voting ensemble'' method to detect the content of the seal title. Five models are utilized in the method, namely Mask R-CNN~\cite{He2017MaskR}, K-Net~\cite{zhang2021knet}, Segformer~\cite{Strudel2021SegmenterTF}, Segmenter~\cite{strudel2021segmenter}, and UperNet~\cite{xiao2018unified}. Each model generates a mask. And they utilize a majority vote to derive the final mask, which allows them to identify the seal title area on the mask.
\end{itemize}

\textbf{2nd ranking method.} 
``China Mobile Cloud Centre" team  are elaborated in detail from the following two perspectives:
\begin{itemize}
    \item 
        Regarding the synthesized data, the team generated a dataset comprising 7000 seals, including circular, elliptical, rectangular, and triangular seals. Additionally, the team addressed the issue of redundant annotation data. Specifically, the annotation process often resulted in unnecessary parts being included on the sides of elliptical and circular seal text. This redundancy could potentially impact the segmentation and subsequent text recognition. To mitigate this, the team developed a correction program that automatically removes the redundant parts by leveraging the geometric properties of the elliptical ring.
    \item 
        The team utilized a single detection model, specifically the VitDet detection part from EVA's (Exploring the Limits of Masked Visual Representation Learning at Scale) framework. The backbone network employed VIT-Giant, while the network head employed Cascade Mask-Rcnn. Considering the small size of the seals, the network input size was adjusted to 320 * 320. To ensure a smooth mask output, the network head's mask output size was increased from 28 * 28 to 56 * 56.
\end{itemize}

\textbf{3rd ranking method.} 
``Institute of Information Engineering, Chinese Academy of Sciences; Mashang Consumer Finance Co., Ltd" team's competition solution is based on the TPSNet detection model~\cite{Wang2021TPSNetRT}. To better adapt the seals, the team has modified the regression branch to regress the bezier control points~\cite{Liu2020ABCNetRS}. Due to some seal titles being too long for a single-stage model to regress, the team has designed a regression-merging post-process where the regressed curves belonging to the same title are merged, weighted by the distance between the feature location and text boundary. The backbone of the team's model is ResNet50 with DCN, pretrained on ImageNet.

In terms of data, the team has designed a script to convert the original polygon annotation to two long curves for every title, allowing training of the regression-based model. Inaccurate or wrong annotations have been re-annotated to ensure data quality. To increase the size of the training set, the team has implemented a seal synthesis pipeline based on Synthtext~\cite{synthtext}. They have modified the character layout to create various seals, used the WTW document dataset images\footnote{https://github.com/wangwen-whu/WTW-Dataset} as background images, and employed the Company-Name-Corpus\footnote{https://github.com/wainshine/Company-Names-Corpus} as the corpus of seal titles. This effort has resulted in the generation of 10,000 synthetic seals that have been added to the training set.

During training and testing, both real and synthetic seals are trained together. The team has applied ColorJitter, Random Rotate, and Random Resize as training augmentations. For testing, an input scale of 448×448 is used without any augmentation.

\subsection{Task 2 End-to-end Seal Title Recognition}
The result for Task 2 is presented on Table~\ref{tab:task2_results}.

\begin{table*}[htbp]
\centering
\caption{Task-2: End-to-end Seal Title Recognition Results.}
\scalebox{0.9}{
\begin{tabular}{lllll}
\hline
Rank & Method Name & Team Members & Insititute & Accuracy  \\
\hline
1    & SPDB LAB                      & \makecell[l]{Jie Li ,Wei Wang,\\Yuqi Zhang, Ruixue Zhang,\\Yiru Zhao, Danya Zhou,\\Di Wang, Dong Xiang,\\Hui Wang, Min Xu,\\Pengyu Chen, Bin Zhang,\\Chao Li, Shiyu Hu,\\Songtao Li, Yunxin Yang} & \makecell[l]{Shanghai \\Pudong \\Development Bank }                                                                    & 91.88\%  \\
\hline
2    & rec320\_3                     & Huajian Zhou                                                                                                                                                   & \makecell[l]{China Mobile \\Cloud Centre  }                                                                          & 91.74\%  \\
\hline
3    & \makecell[l]{Dao Xianghu\\ light of TianQuan} &\makecell[l] {Kai Yang, Ye Wang,\\ Bin Wang, Wentao Liu, \\Xiaolu Ding, Jun Zhu, \\Ming Chen, Peng Yao, \\Zhixin Qiu}                                                                 & \makecell[l]{CCB \\Financial Technology \\Co. Ltd, China}                                                              & 91.22\%  \\
\hline
4    & AppAI for Seal                & \makecell[l]{Chuanjian Liu, Miao Rang,\\ Zhenni Bi, Zhicheng Liu,\\ Wenhui Dong, Yuyang Li,\\ Dehua Zheng, Hailin Wu,\\ Kai Han, Yunhe Wang}                                         & Noah                                                                                                 & 90.20\%  \\
\hline
5    & ensemble                      & xubo                                                                                                                                                           & -                                                                                                    & 90.08\%  \\
\hline
6    & task2\_test\_submit2          & jgj aksbob                                                                                                                                                     & pa                                                                                                   & 88.90\%  \\
\hline
7    & INTIME\_OCR(e2e)              &\makecell[l] {Wei Wang, Jin Wei, \\Chengxiang Ran, Xinye Yang,\\ Tianjiao Cao, Fangmin Zhao}                                                                                      & \makecell[l]{Institute of \\Information Engineering,\\ Chinese Academy \\of Sciences;\\ Mashang Consumer\\ Finance Co., Ltd} & 84.24\%  \\
\hline
8    & task2 result                  & DH                                                                                                                                                             & -                                                                                                    & 84.22\%  \\
\hline
9    & Seal Recognize                & \makecell[l]{Shente Zhou, Tianyi Zhu, \\Weihua Cao, Mingchao Fang, \\Xiaogang Ouyang}                                                                                                & Shizai Intellect                                                                                     & 83.02\%  \\
\hline
10   & SealRecognizor                & Qiao Liang                                                                                                                                                     & Zhejiang University                                                                                  & 83.00\%\\

\hline
\end{tabular}}
\label{tab:task2_results}
\end{table*}

The methods used by the top 3 submissions for Task 1 are
presented below.

\textbf{1st ranking method.} 
``Shanghai Pudong Development Bank" team's method can be described in detail from the following two perspectives:
\begin{itemize}
    \item Circle seals and Ellipse seals:
        Based on the results of the circle and ellipse seals title detection in task1, PCA technology was used to correct the rotated seal, the image processing technology was used to separate the seal title,  and finally the curved text was sent to the recognition model for recognition. The recognition model was selected by Trocr~\cite{Li2021TrOCRTO}, and the training data includes the provided training data and synthetic data.
    \item Rectangle seals and Triangle seals:
        Rectangle seals and triangle seals were not based on the task1 detection model, but train a text line detection mode~\cite{Wang2019EfficientAA}l. the image processing technology was used to separate the seal title. The recognition model was selected by Trocr~\cite{Li2021TrOCRTO}, and the training data includes the provided by synthetic data.
\end{itemize}

\textbf{2nd ranking method.} 
``China Mobile Cloud Centre"  team's method can be described in detail from the following perspectives:

\begin{itemize}
    \item Text Detection and Segmentation Module:

    Data:

    \begin{enumerate}
    \item Synthesize data: Synthesize 7000 seals (including circular, elliptical, rectangular, and triangular seals).
    
    \item Correction of annotation data: In the annotation process, there are redundant parts at two sides of elliptical and circular seal text, which can affect text recognition. The correct program removes the redundant parts automatically by utilizing the geometric properties of the elliptical ring.
    \end{enumerate}
    
    Method:
    
    The text detection and segmentation module follows the following approach:
    
    \begin{enumerate}
    \item Detection Model: Only one detection model is used, using EVA's (Exploring the Limits of Masked Visual Representation Learning at Scale) VitDet detection part.

    \item Backbone Network: The backbone network uses VIT-Giant.
    
    \item Network Head: The network head uses Cascade Mask-Rcnn.
    
    \item Adjustment for Seal Size: Considering the small size of the seals, the network input size is adjusted to $320 \times 320$.
    
    \item Mask Output Enhancement: The network head's mask output size is adjusted from $28 \times 28$ to $56 \times 56$ to achieve a smooth mask output.
    \end{enumerate}

    \item Text Rectification Module:

    \begin{enumerate}
    \item For triangular and rectangular texts, the text is rectified to horizontal text by using the direction classification model combined with affine transformation.

    \item For elliptical and circular text, the least squares method is used to obtain the upper and lower curve equations of the text. Based on the curve equations, the curve region is divided into several small regions, and affine transformations are performed on these regions. Then, they are concatenated to get the horizontal text.
    \end{enumerate}

    \item Text Recognition Model:

    Data:
    
    \begin{enumerate}
    
    \item Synthetic data: Extracting millions of lines of corpus from the open-source THUCNews, News2016zh, and wiki\_zh\ corpus datasets, and using this data to synthesize horizontal and curved text images.
    
    \item Rectification data: Rectifying or cropping official training images to obtain text images.
    \end{enumerate}
    
    Method:
    
    \begin{enumerate}
    \item Recognition model 1: Using SVTR-Small (Scene Text Recognition with a Single Visual Model), with the network input size adjusted to $48 \times 320$.
    
    \item Recognition model 2: Using DIG (Reading and Writing: Discriminative and Generative Modeling for Self-Supervised Text Recognition), with the network input size adjusted to $48 \times 288$.
    
    \item Recognition correction: If the results of the two models are different, the correction program uses the open-source Chinese administrative division dataset for correction.
    \end{enumerate}

\end{itemize}

\textbf{3rd ranking method.} 
``CCB Financial Technology Co. Ltd, China" team finds that the difficulties of recognition mainly focus on multi-directional recognition, overlapping interference from handwritten or printed characters, fuzzy and blurred images, and multiple reading orders, after data exploration and analysis. Based on the analysis, they build the following solution. First, they make a seal title segmentation that masks out the non-title area, and removes the interference of irrelevant regions. Then, they train a TrOCR~\cite{Li2021TrOCRTO} model using over 6 million data from the training set, open dataset, and synthetic dataset. Finally, in the post process, place names correction is implemented.

In the seal title segmentation, they adopt an ensemble strategy with five segmentation models to vote for the title segmentation, laying a good foundation for the recognition.
Since the training set only has 5000 images, it is far from enough for the recognition task. They use the official chars.txt dictionary and collect the corpus of company names and organization names on the Internet, and generate a large number of seals by codes. To simulate the real situation, they use various fonts, colors, backgrounds, and textures to synthesize the images, and they perform kinds of data augmentation strategies for improving generalization including rotation, gaussian blur, stretching, perspective transformation, contour expansion or contraction and so on. In addition, they use 10k seals from Baidu public dataset.

At the early stage of the competition, they use the public dataset and the synthesized dataset as the training set and the original training set of the competition as the test set. They continuously synthesize kinds of data to improve the accuracy of the test set.
To further improve the accuracy, they design a classifier to separate circular seals (Circle/Ellipse shapes) and non-circular seals (Rectangle/Triangle shapes). They generate nearly 400k non-circular seals. And they compare the single recognition model solution with the solution of classifying then recognizing with multiple models. And they verify that the former solution is better.

When analyzing bad cases, they find that smudging and character overlapping often lead to recognition errors. So they design place names based post-processing strategy to correct some of these errors.

\section{Discussion}
\textbf{Task 1 Seal Title Text Detection.} Many teams use data augmentation techniques such as random scaling, flipping, rotation, cropping, and synthesis of hard samples with various shapes, colors, fonts, and backgrounds to improve the generalization ability of their models. Additionally, they employ powerful methods such as DBNet++~\cite{liao2022real}, ResNet, Segformer, Unet, and PANNet to enhance performance. Some teams also use a ``voting ensemble" strategy to achieve better results. Certain teams take into account the four main shapes of seal images during training and data generation. Task 1, which involves general text detection, has produced excellent results, with 12 teams achieving an Hmean of over 95\%. However, in scenarios that require stricter standards or zero error rates, even the best method with a 99.06\% Hmean in Task 1 cannot meet the required performance. Therefore, there is still room for improvement, and further efforts and exploration are needed.

\textbf{Task 2 End-to-end Seal Title Recognition.} 
Extending from Task 1, the task of end-to-end seal title recognition poses greater challenges, requiring flexible adjustments and further refinement based on the findings from Task 1. To address the issues of curved or overlapped text, some teams have designed image processing technologies such as PCA or post-processing strategies to correct these errors. Diverse model ensembles continue to be utilized for improved results. For more accurate recognition, powerful models such as Parseq~\cite{bautista2022parseq} and TrOCR are employed. To further enhance accuracy, many teams have designed a novel classification network to differentiate circular seals (Circle/Ellipse shapes) and non-circular seals (Rectangle/Triangle shapes). Additionally, pre-trained models and fine-tuning strategies with augmentations and label smoothing based on joint datasets, including training sets, open datasets such as Synthtext-Chinese, ReCTs~\cite{Liu2019ICDAR2R}, LSVT~\cite{Sun2019ICDAR2C}, ArT~\cite{Chng2019ICDAR2019RR}, and synthetic datasets, are used. However, as shown in Table~\ref{tab:task2_results}, only five teams achieved accuracy of over 90.00\%, with the top-1 accuracy being 91.88\%. Therefore, we can conclude that end-to-end seal title recognition remains a challenging task, with most methods submitted using different ideas and approaches. We look forward to seeing more innovative approaches proposed following this competition.

\section{Conclusion}
We organized the ReST competition, with a focus on reading challenging seal title text, an area that has not received sufficient attention in the field of document analysis. To this end, we constructed new datasets, comprising 10,000 real seal data labeled with text polygons and transcripts. Strong interest from both academia and industry was evident, with a large number of submissions showcasing novel ideas and approaches for the competition tasks. Reading seal titles holds significant potential for numerous document analysis applications, making it a rewarding task. However, despite the top-performing team achieving a remarkable performance of approximately 99\% in Task 1, the task still warrants continued research and exploration, particularly in strict scenarios with zero error tolerance rates. Additionally, Task 2 proved to be a challenging task, with only five teams achieving an accuracy above 90\%, and the top-1 accuracy reaching 91.88\%. However, the excellent submissions by these teams provide valuable insights for other researchers. Future competitions could expand on this topic with more challenging datasets and applications, thus attracting researchers from the fields of computer vision and advancing the state-of-the-art in document analysis.

\section{Acknowledgements} This competition is supported by the National Natural Science Foundation of China (No.62225603, No.62206103, No.62206104). The organizers thank Sergi Robles and the RRC web team for their tremendous support on the registration, submission and evaluation jobs.
\bibliographystyle{splncs04}
\bibliography{main}
\end{document}